\documentclass{article}

\usepackage[utf8]{inputenc} % allow utf-8 input
\usepackage[T1]{fontenc}    % use 8-bit T1 fonts
\usepackage[ruled,vlined,lined,noend]{algorithm2e}
\usepackage{url}            % simple URL typesetting
\usepackage{amsfonts}       % blackboard math symbols
\usepackage{wrapfig}
\usepackage{microtype}
\usepackage{booktabs}

\usepackage{amsmath}
\usepackage{epsfig}
\usepackage{graphicx}
\usepackage{color}
\usepackage{siunitx}
\usepackage{bm}
\usepackage{multirow}
\usepackage{hyperref}
\usepackage{caption}
\usepackage{subcaption}

\usepackage[accepted]{icml2020}

\newcommand{\argmax}{\operatornamewithlimits{arg\ max}}

\icmltitlerunning{Occlusion resistant learning of intuitive physics from videos}

\begin{document}

\twocolumn[
\icmltitle{Occlusion resistant learning of intuitive physics from videos}

\begin{icmlauthorlist}
 \icmlauthor{Ronan Riochet}{ens,inria}
 \icmlauthor{Josef Sivic}{ens,inria,prague}
 \icmlauthor{Ivan Laptev}{ens,inria}
 \icmlauthor{Emmanuel Dupoux}{ens,inria,fb}
\end{icmlauthorlist}

\icmlaffiliation{ens}{Ecole Normale Supérieure, CNRS, PSL Research University, Paris, France.}
\icmlaffiliation{inria}{Inria, Paris.}
\icmlaffiliation{fb}{Facebook AI Research}
\icmlaffiliation{prague}{Czech Institute of Informatics, Robotics and Cybernetics, Czech Technical University in Prague.}

\icmlcorrespondingauthor{Ronan Riochet}{ronan.riochet@polytechnique.edu}
\vskip 0.3in
]

\printAffiliationsAndNotice

\begin{abstract}
To reach human performance on complex tasks, a key ability for artificial systems is to understand physical interactions between objects, and predict future outcomes of a situation. This ability, often referred to as \textit{intuitive physics}, has recently received attention and several methods were proposed to learn these physical rules from video sequences. Yet, most of these methods are restricted to the case where no, or only limited, occlusions occur. 
In this work we  propose  a  probabilistic  formulation of learning intuitive physics in 3D scenes with significant inter-object occlusions. In our formulation, object positions are modelled as latent variables enabling the reconstruction of the scene. 
We then propose a series of approximations that make this problem tractable. 
Object proposals are linked across frames using a combination of a recurrent interaction network, modeling the physics in object space, and a compositional renderer, modeling the way in which objects project onto pixel space.
We demonstrate significant improvements over state-of-the-art in the intuitive physics benchmark of~\citet{riochet_intphys:_2018}. We apply our method to a second dataset with increasing levels of occlusions, showing it realistically predicts segmentation masks up to 30 frames in the future. Finally, we also show results on predicting motion of objects in real videos.

\end{abstract}

\section{Introduction}

Learning intuitive physics has recently raised significant interest in the machine learning literature. To reach human performance on complex visual tasks, artificial systems need to understand the world in terms of macroscopic objects, movements, interactions, etc. Infant development experiments show that young infants quickly acquire an intuitive grasp of how objects interact in the world, and that they use these intuitions for prediction and action planning \cite{carey_origin_2009,baillargeon_core_2012}. This includes the notions of gravity \cite{carey_origin_2009}, continuity of trajectories \cite{spelke_spatiotemporal_1995}, collisions \cite{saxe_perception_2006}, etc. Object permanence, the fact that an object continues to exist when it is occluded, \cite{kellman_perception_1983}, is one of the first concepts developed by infants.

From a modeling point of view, the key scientific question is how to develop general-purpose methods that can make physical predictions in noisy environments, where many variables of the system are unknown.
A model that could mimic even some of infant's ability to predict the dynamics of objects and their interactions would be a significant advancement in model-based action planning for robotics \cite{agrawal_learning_2016,finn_deep_2017}. The laws of macroscopic physics are relatively simple and can be readily learned when formulated in 3D cartesian coordinates \cite{battaglia_interaction_2016,mrowca_flexible_2018}. 

However, learning such laws from real world scenes are difficult for at least two reasons. First, estimating accurate 3D position and velocity of objects is challenging when only their retinal projection is known, even assuming depth information, because partial occlusions by other objects render these positions ambiguous. Second, objects can be fully occluded by other objects for a significant number of frames. 

In this paper we address these issues and develop a model for learning intuitive physics in 3D scenes with significant inter-object occlusions. We propose a probabilistic formulation of the intuitive physics problem, whereby object positions are modelled as latent variables enabling the reconstruction of the scene. We then propose a series of approximations that make this problem tractable. 
In detail, proposals of object positions and velocities (called object states) are derived from object masks, and then linked across frames using a combination of a recurrent interaction network, modeling the physics in object space, and a compositional renderer, modeling the way in which objects project onto pixel space.

Using the proposed approach, we show that it is possible to follow object dynamics in 3D environments with severe inter-object occlusions. We evaluate this ability on the IntPhys benchmark \cite{riochet_intphys:_2018}, a benchmark centered on classifying videos as being physically possible or not. We show better performance compared to \cite{riochet_intphys:_2018,Smith2019ModelingEV}. A second set of experiments show that it is possible to learn the physical prediction component of the model even in the presence of severe occlusion, and predict segmentation masks up to 30 frames in the future. Ablation studies and baselines \cite{battaglia_interaction_2016} evaluate the importance of each component of the model, as well the impact of occlusions on performance. 

Our model is fully compositional and handles variable number of objects in the scene. Moreover, it does not require as input (or target) annotated inter-frame correspondences  during training. Finally, our method still works with no access to ground-truth segmentation, using (noisy) outputs from a pre-trained object/mask detector~\cite{he_mask_2018}, a first step towards using such models on real videos.

\section{Related work}

\paragraph{Forward modeling in videos.}
Forward modeling in videos has been studied for action planning \cite{ebert_visual_2018,finn_unsupervised_2016} and as a scheme for unsupervised learning of visual features~\cite{lan_hierarchical_2014,mathieu_deep_2015}. In that setup, a model is given a sequence of frames and generates frames in future time steps  \cite{lan_hierarchical_2014,mathieu_deep_2015,finn_unsupervised_2016,DBLP:journals/corr/abs-1806-04768,zhu_object-oriented_2018,DBLP:journals/corr/VillegasYHLL17,DBLP:journals/corr/abs-1906-07889}. Such models tend to perform worse when the number of objects increases, sometimes failing to preserve object properties and generating blurry outputs.

\paragraph{Learning dynamics of objects.}
Longer term predictions can be more successful when done on the level of trajectories of individual objects. For example, in \cite{wu_neural_2017}, the authors propose "scene de-rendering", a system that builds an object-based, structured representation from a static (synthetic) image. The recovered state can be further used for physical reasoning and future prediction using an off-the-shelf physics engine on both synthetic and real data~\cite{battaglia_simulation_2013,wu_learning_2017,Smith2019ModelingEV,DBLP:journals/corr/abs-1906-03853}. Future prediction from static image is often multi-modal (e.g. car can move forward or backward) and hence models able to predict  multiple possible future predictions, e.g. based on variational auto-encoders~\cite{xue_visual_2016}, are needed. Autoencoders have been also applied to learn the dynamics of video \cite{kosiorek_sequential_2018,hsieh_learning_nodate} in restricted 2D set-ups and/or with a limited number of objects.

Others have developed structured models that factor object motion and object rendering into two learnable modules. 
Examples include \cite{watters_visual_2017,marco_fraccaro_disentangled_2017,ehrhardt_learning_2017,ehrhardt_taking_2017} that combine object-centric dynamic models and visual encoders. Such models parse each frame into a set of object state representations, which are used as input of a "dynamic" model, predicting object motion. However, \cite{marco_fraccaro_disentangled_2017} restrict drastically the complexity of the visual input by working on binary 32x32 frames, and \cite{ehrhardt_learning_2017,ehrhardt_taking_2017,watters_visual_2017} still need ground truth position of objects as input or target~\cite{watters_visual_2017} for training.
However, modeling 3D scenes with significant inter-object occlusions, which is the focus of our work, still remains an open problem.  

In our work, we build on learnable models of object dynamics \cite{battaglia_interaction_2016} and \cite{chang_compositional_2016}, which have the key property that they are compositional and hence can model a variable number of objects, but extend them to learn from visual input rather than ground truth object state vectors.

Our work is also related to~\cite{janner_reasoning_2018}, who combine an object-centric model of dynamics with a differentiable renderer to predict a single image in a future time, given a single still image as input. In contrast, we develop a probabilistic formulation of intuitive physics that (i) predicts the physical plausibility of an observed dynamic scene, and (ii) infers velocities of objects as latent variables allowing us to predict full trajectories of objects through time despite long complete occlusions. 
Others have proposed unsupervised methods to discover objects and their interactions in 2D videos~\cite{van_steenkiste_relational_2018}.  
It is also possible to construct Hierarchical Relation Networks~\cite{mrowca_flexible_2018} or particle-based models \cite{DBLP:journals/corr/abs-1810-01566}, representing objects as graphs and predicting interactions between pairs of objects. However, this task is still challenging and requires full supervision in the form of ground truth position and velocity of objects.

\paragraph{Learning physical properties from visual inputs.}
Related are also methods for learning physical representations from visual inputs. Examples include  \cite{greff_multi-object_2019,DBLP:journals/corr/abs-1901-11390} who focus on segmenting images into interpretable objects with disentangled representations.
Learning of physical properties, such as mass, volume or coefficients of friction and restitution, has been considered in~\cite{wu_physics_2016}. Others have looked at predicting the stability and/or the dynamics of towers of blocks \cite{lerer_learning_2016,zhang_comparative_2016,li_fall_2016,li_visual_2016,mirza_generalizable_2017,groth2018shapestacks}. 
Our work is complementary. We don't consider prediction of physical properties but focus on learning models of object dynamics handling inter-object occlusions at both training and test time.

\begin{figure}[t!]
\begin{center}
\includegraphics[width=0.99\linewidth]{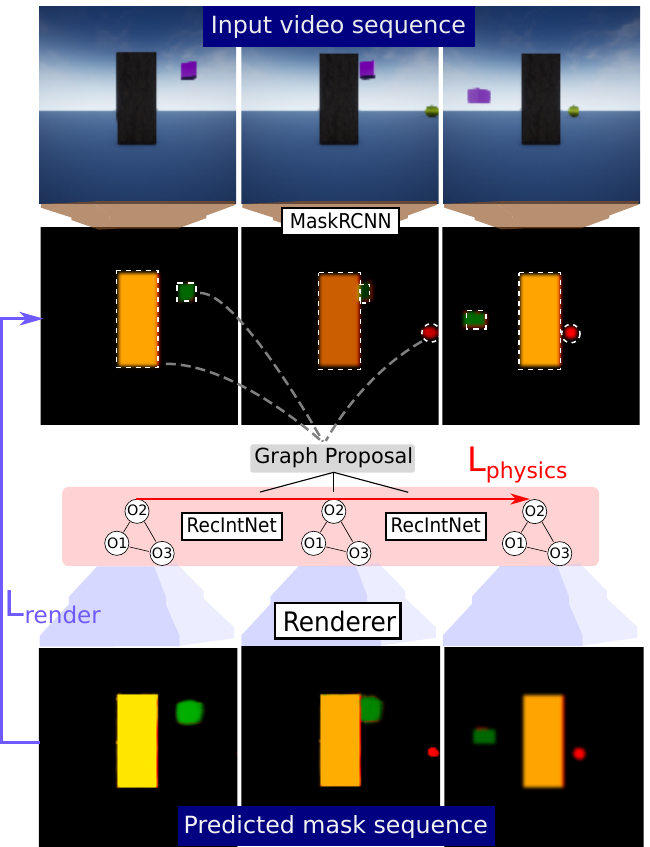}
\end{center}
\caption{\small {\bf Overview of our occlusion resistant intuitive physics model.} A pre-trained object detector (MaskRCNN) returns object detections and masks (top). A \textit{graph proposal} matching links object proposals through time: from a pair of frames the Recurrent Interaction Network ($RecIntNet$) predicts next object position and matches it with the closest object proposal. If an object disappears (e.g. due to occlusion - no object proposal), the model keeps the prediction as an \textit{object state}, otherwise this object state is updated with the observation. Finally, the Compositional Rendering Network ($Renderer$) predicts masks from object states and compares them with the observed masks. The errors of predictions of $RecIntNet$ and $Renderer$ on the full sequence are summed into a \textit{physics} and a \textit{render} loss, respectively. The two losses are used to assess whether the observed scene is physically plausibility.}
\label{fig:model}
\end{figure}

\section{Occlusion resistant intuitive physics}
This section describes our model for occlusion resistant learning of intuitive physics. In section~\ref{sec:overview} we present an overview of the method, then describe it's two main components: the occlusion-aware compositional renderer that predicts object masks given a scene state representation (section~\ref{sec:rend}), and the recurrent interaction network that predicts the scene state evolution over time (section~\ref{sec:dyn}). Finally, in section~\ref{sec:event} we describe how these two components are used jointly to decode an entire video clip.

\subsection{Intuitive physics via event decoding}
\label{sec:overview}
We formulate the problem of \textit{event decoding} as that of assigning to a sequence of video frames  
$F=f_{t=1..T}$ a sequence of underlying object states $S=s^{i=1..N}_{t=1..T}$ that can explain (i.e. reconstruct) this sequence of frames. By object state, we mean object positions, velocities and categories. 
Within a generative probabilistic model, we therefore try to find the state $\hat{S}$ that maximizes $P(S|F,\theta)$, where $\theta$ is a parameter of the model: $\hat{S}= \argmax_{S} P(S|F,\theta)$. A nice property of this formulation is that we can use $P(\hat{S}|F,\theta)$ as a measure of the \textit{plausibility} of a video sequence, which is exactly the metric required in the Intphys benchmark. 

With Bayes rule, $P(S|F,\theta)$ decomposes into the product of two probabilities that are easier to compute, $P(F|S,\theta)$, the \textit{rendering model}, and $P(S|\theta)$, the \textit{physical model}. This is similar to the decomposition into an acoustic model and a language model in ASR \cite{jelinek1997}. The event decoding problem then becomes:

\vspace{-2.0em}
\begin{equation}
\label{eq:intepretation}
\begin{split}
\hat{S}&= \argmax_{S} P(F|S,\theta) P(S|\theta).
\end{split}
\end{equation}
\vspace{-2.0em}

Such a formulation naturally accounts for occlusion through the rendering model which maps underlying positions into the visible outcome in pixel space. During inference, the physical model is used to fill in the blanks, i.e., imagine what happens behind occluders to maximize the probability of the trajectory. As for the learning problem, it can be formulated as follows:

\vspace{-1.5em}
\begin{equation}
\label{eq:learning}
\begin{split}
\hat{\theta}&= \argmax_{\theta} P(F|\theta).
\end{split}
\end{equation}
\vspace{-2.0em}

In this paper we will apply a number of simplifications to make this problem tractable. First, we operate in mask space and not in pixel space. This is done by using an off-the shelf instance mask detector (Mask-RCNN~\cite{he_mask_2018}), making the task of rendering easier, since all of the details and  textures are removed from the reconstruction problem. Therefore $F$ is a sequence of (stacks of) binary masks for different objects in the scene. Second, the state space is expressed, not in 3D coordinates, which would require to learn inverse projective geometry, but directly in retinotopic pixel coordinate plus depth (2.5D, something easily available in RGBD cameras). It turns out that learning physics in this space is not more difficult than in the true 3D space. Finally, the probabilistic models are implemented as Neural Networks. The rendering model ($Renderer$) is implemented as a neural network mapping object states into pixel space. The physical model is implemented as a recurrent interaction network ($RecIntNet$), mapping object state at time $t$ as a function of past states.

In practice, computing the $\argmax$ in eq.~\eqref{eq:intepretation} is difficult because the states are continuous, the number of objects is unknown, and some objects are occluded in certain frames, yielding a combinatorial explosion regarding how to link hypothetical object states across frames. In this paper, we propose a major approximation to help solving this problem by proceeding in two steps. In the first step, a \textit{scene graph proposal} is computed using bounding boxes to estimate object position, nearest neighbor matching across nearby frames to estimate velocities, and the roll-out of the physics engine to link the objects across the entire sequence (which is critical to deal with occlusions). The second step consists of optimizing $S$ (given by eq.~\eqref{eq:intepretation}) by using gradient descent on both models, capitalizing on the fact that both models are differentiable. More precisely, rather than computing probabilities explicitly, we define two losses (that can be interpreted as a proxy for negative log probability): (i) the rendering loss $L_\text{render}$ that measures the discrepancy between the masks predicted by the renderer and the observed masks in individual frames; and (ii) the physical loss $L_\text{physics}$ that measures the discrepancy between states predicted by the recurrent interaction network ($RecIntNet$) and the actual observed states.  As in ASR, we will combine these two losses with a scaling factor $\lambda$, yielding a total loss: 
\vspace{-1.5em}
\begin{equation}
\begin{split}
L_\text{render}(S,F)&= \sum_{t=1}^{T}L_{\text{mask}}({Renderer}(s_t), F),\\
L_\text{physics}(S)&=\sum_{t=1}^{T-1}\|s_{t+1}-{RecIntNet}(s_t)\|^2,\\
L_\text{total}(S,F)&= \lambda L_\text{render}(S,F) + (1 - \lambda) L_\text{physics}(S).
\end{split}
\end{equation}
\vspace{-1.0em}

$L_\text{mask}$ is a pixel-wise loss defined in detail in the supplementary material.

We use the total loss as the objective function to minimize in order to find the interpretation $\hat{S}$ of the masks of a video clip $F$. And it will be used to provide a plausibility score to decide whether a given scene is physically plausible in the evaluation on the IntPhys Benchmark (section~\ref{sec:intphys}). As for learning, instead of marginalizing over possible state, we will just optimize the parameters over the point estimate optimal state $\hat{S}$. The aim of this paper is to show that these approximations notwithstanding, a system constructed according to this set-up can yield good results.

\subsection{The Compositional Renderer}\label{sec:rend}

\begin{figure}[tbp]
\begin{center}
\includegraphics[width=0.99\linewidth]{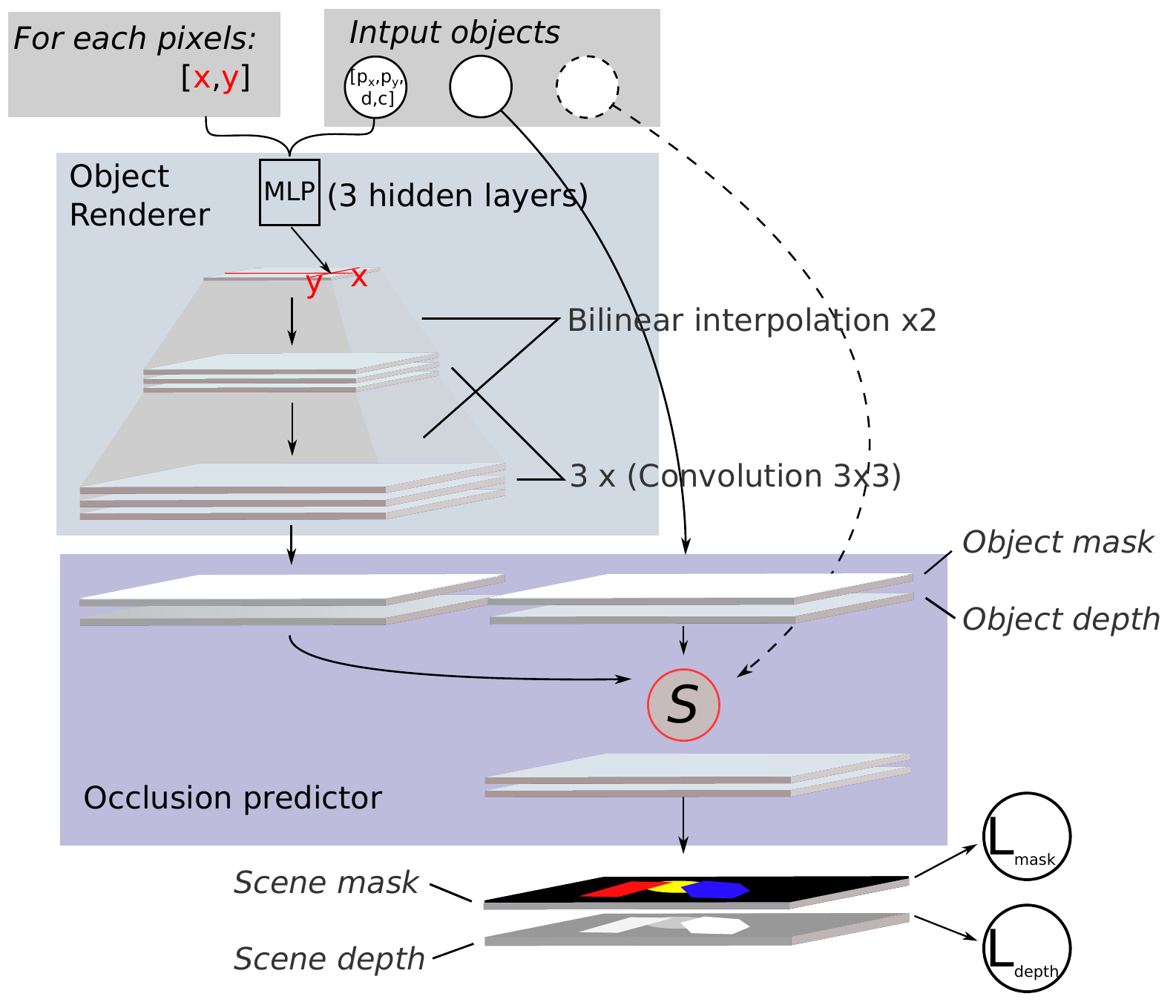}
\end{center}
\caption{\small {\bf Compositional Rendering Network ($Renderer$)} Takes as input a list of object states. First, the {\em object rendering network} reconstructs a segmentation mask and a depth map for each object independently. Second,  the {\em occlusion predictor} composes all predicted object masks into the final scene mask, generating the appropriate pattern of inter-object occlusions obtained from the predicted depth maps of the individual objects.}
\label{fig:renderer}
\end{figure}

We introduce a differentiable \textit{Compositional Rendering Network} (or $Renderer$) that predicts a segmentation mask in the image given a list of $N$ objects specified by their $x$ and $y$ position in the image, depth and possibly additional properties such as object type (e.g. sphere, square, ...) or size. Importantly, our neural rendering model has the ability to take a variable number of objects as input and is invariant to the order of objects in the input list. It contains two modules (see Figure~\ref{fig:renderer}). First, the {\em object rendering network} reconstructs a segmentation mask and a depth map for each object. Second,  the {\em occlusion predictor} composes the $N$ predicted object masks into the final scene mask, generating the appropriate pattern of inter-object occlusions obtained from the predicted depth maps of the individual objects.

\paragraph{The Object rendering network}
takes as input a vector of $l$ values corresponding to the position coordinates $(x^k, y^k, d^k)$ of object $k$ in a frame together with additional dimensions for intrinsic object properties (shape, color and size) $(\bm{c})$. The network predicts object's binary mask, $M^k$ as well as the depth map $D^k$.  The input vector $(x^k, y^k, d^k, \bm{c}^k) \in \mathbb{R}^l$ is first copied into a $(l+2)\times 16 \times 16$ tensor, where each $16\times16$ cell position contains an identical copy of the input vector together with $x$ and $y$ coordinates of the cell. Adding the $x$ and $y$ coordinates may seem redundant, but this kind of \textit{position field} enables a very local computation of the shape of the object and avoids a large number of network parameters (similar architectures were recently also studied in \cite{DBLP:journals/corr/abs-1807-03247}).

The input tensor is processed with $1\times 1$ convolution filters. The resulting 16-channel feature map is further processed by three blocks of convolutions. Each block contains three convolutions with filters of size $1\times 1$, $3 \times 3$ and $1 \times 1$ respectively, and  $4$, $4$ and $16$ feature maps, respectively. We use \texttt{ReLU} pre-activation before each convolution, and up-sample (scale of $2$ and bilinear interpolation) feature maps between blocks. The last convolution outputs $N+1$ feature maps of size $128\times 128$, the first feature map encoding depth and the $N$ last feature maps encoding mask predictions for the individual objects. The object rendering network is applied to all objects present, resulting in a set of masks and depth maps denoted as $\{(\hat{M}^k, \hat{D}^k), k=1..N\}$.

\paragraph{The Occlusion predictor} takes as input the masks and depth maps for $N$ objects and aggregates them to construct the final occlusion-consistent mask and depth map. To do so it computes, for each pixel $i,j\leq128$ and object $k$ the following weight: 

\vspace{-2em}
\begin{align}
\label{eq:softmin}
    c_{i,j}^k = \frac{e^{\lambda \hat{D}_{i,j}^k}}{\sum_{q=1}^N e^{\lambda \hat{D}_{i,j}^q}}, k=1..N, 
    %\vspace*{-45mm}
\end{align}
where $\lambda$ is a parameter learned by the model. The final masks and depth maps are computed as a weighted combination of masks $\hat{M}_{i,j}^k$  and depth maps $\hat{D}_{i,j}^k$ for individual objects $k$: 
$\hat{M}_{i,j} = \sum_{k=1}^N c_{i,j}^k \hat{M}_{i,j}^k$, $\hat{D}_{i,j} = \sum_{k=1}^N c_{i,j}^k \hat{D}_{i,j}^k$, where $i,j$ are output pixel coordinates $\forall i,j \leq 128$ and $c_{i,j}^k$ the weights given by~\eqref{eq:softmin}. 
The intuition is that the occlusion renderer constructs the final output $(\hat{M},\hat{D})$ by selecting, for every pixel, the mask with minimal depth (corresponding to the object occluding all other objects). For negative values of $\lambda$ equation~\eqref{eq:softmin} is as a softmin, that selects for every pixel the object with minimal predicted depth. Because $\lambda$ is a trainable parameter, gradient descent forces it to take large negative values, ensuring good occlusion predictions.  Also note that this model does not require to be supervised by the depth field to predict occlusions correctly. In this case, the object rendering network still predicts a feature map $\hat{D}$ that is not equal to the depth anymore but is rather an abstract quantity that preserves the relative order of objects in the view. This allows $Renderer$ to predict occlusions when the target masks are RGB only. However, it still needs depth information in its input (true depth or rank order).

\begin{figure}[h]
\begin{center}
\includegraphics[width=1\linewidth]{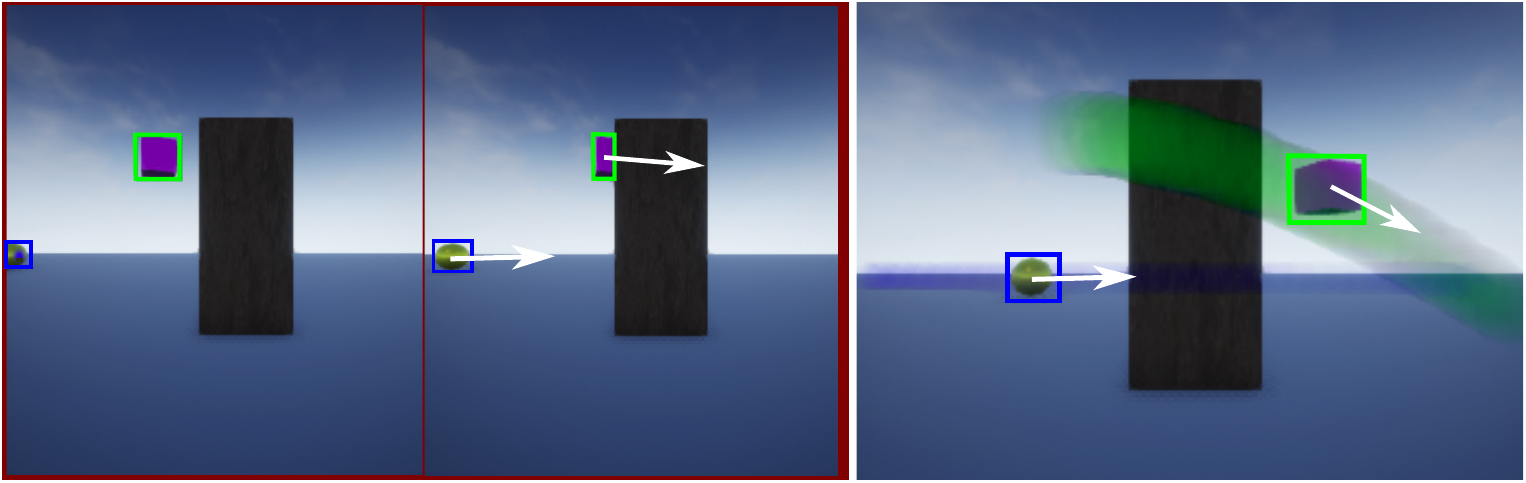}
\end{center}
\caption{
{\bf Illustration of event decoding in the videos of the IntPhys dataset.} A pre-trained object detector returns object proposals  in the video (bounding boxes). An initial match is made across two seed neighbouring frames, also estimating object velocity (left, white arrows). The dynamic model (RecIntNet) predicts object positions and velocities in future frames, enabling the match of objects despite significant occlusions (right, bounding box colors and highlights).}
\label{fig:illustration}
\end{figure}

\subsection{The Recurrent Interaction Network ($RecIntNet$)}
\label{sec:dyn}
To model object dynamics, we build on the Interaction Network~\cite{battaglia_interaction_2016}, which predicts dynamics of a variable number of objects by modeling their pairwise interactions. 
Here we describe three extensions of the vanilla Interaction Network model. First, we extend the Interaction Network to model 2.5D scenes where position and velocity have a depth component. Second, we turn the Interaction Network into a recurrent network. Third, we introduce variance in the position predictions, to stabilise the learning phase, and avoid penalizing too much very uncertain predictions. The three extensions are described below. 

\paragraph{Modeling compositional object dynamics in 2.5D scenes.}
As shown in \cite{battaglia_interaction_2016}, Interaction Networks can be used to predict object motion both in 3D or in 2D space. Given a list of objects represented by their positions, velocities and size in the Cartesian plane, an Interaction Network models interactions between all pairs of objects, aggregates them over the image and predicts the resulting motion for each object. Here, we model object interactions in 2.5D space, since we have no access to the object position and velocity in the Cartesian space. Instead we have locations and velocities in the image plane plus depth (the distance between the objects and the camera).

\paragraph{Modeling frame sequences.} 
The vanilla Interaction Network~\cite{battaglia_interaction_2016} is trained to predict position and velocity of each object in one step into the future. Here, we learn from multiple future frames. We "rollout" the Interaction Network to predict a whole sequence of future states as if a standard Interaction Network was applied in recurrent manner. We found that faster training can be achieved by directly predicting changes in the velocity, hence:

\vspace{-2em}
\label{eq:taylor}
\begin{align}
[p_1, v_1, c] = [p_0 + \delta t v_0 +  \frac{\delta t^2}{2} \mathbf{d_v}, v_0 + \mathbf{d_v}, c],
\end{align}
where $p_1$ and $v_1$ are position and velocity of the object at time $t_1$, $p_0$ and $v_0$ are position and velocity at time $t_0$, and $\delta t = t_1 - t_0$ is the time step. Position and velocity in pixel space ($p=[p_x,p_y,d]$ where $p_x,p_y$ are the position of the object in the frame), $d$ is depth and $v$ is the velocity in that space. Hence $\mathbf{d_v}$ can be seen as the \textit{acceleration}, and $(v_0 + \mathbf{d_v})$,$(p_0 + \delta t v_0 +  \frac{\delta t^2}{2} \mathbf{d_v})$ as the first and second order Taylor approximations of velocity and position, respectively.
Assuming an initial weight distribution close to zero, this gives the model a prior that the object motion is linear.

\paragraph{Prediction uncertainty.}

To account for prediction uncertainty and stabilize learning, we assume that object position follows a multivariate normal distribution, with diagonal covariance matrix. Each term $\sigma_x^2$, $\sigma_y^2$, $\sigma_d^2$ of the covariance matrix represents the uncertainty in prediction, along x-axis, y-axis and depth. Such uncertainty is also given as input to the model, to account for uncertainty either in object detection (first prediction step) or in the recurrent object state prediction. The resulting loss is negative log-likelihood of the target $p_1$ w.r.t. the multivariate normal distribution, which reduces to: %

\vspace*{-2em}
\begin{equation}
    \label{eq:logvar}
    \mathcal{L}\big((\hat{p_1}, \hat{\tau_1}), p_1\big) = \frac{(\hat{p_1}-p_1)^2}{\exp \hat{\tau_1}} + \hat{\tau_1},
\end{equation}

 where $\hat{\tau_1}=\log\hat{\sigma^2_1}$ is the estimated level of noise propagated through the Recurrent Interaction Network, where $\sigma_1$ concatenates $\sigma_x^2$, $\sigma_y^2$, $\sigma_d^2$, $p_1$ is the ground truth state and $\hat{p_1}$ is the predicted state at time $t+1$.  
 The intuition is that the squared error term in the numerator is weighted by the estimated level of noise $\hat{\tau_1}$, which acts also as an additional regularizer. We found that modeling the prediction uncertainty is important for dealing with longer occlusions, which is the focus of this work.

\subsection{Event decoding}\label{sec:event}
Given these components, event decoding is obtained in two steps. First, scene graph proposal gives initial values for object states based on visible objects detected on a frame-by-frame basis. These proposed object states are linked across frames using $RecIntNet$ and a nearest neighbor strategy. Second, this initial proposal of the scene interpretation is then optimized by minimizing the total loss by gradient descent through both $RecIntNet$ and $Renderer$ on the entire sequence of object states, yielding the final interpretation of the scene (example in Figure ~\ref{fig:illustration}), as well as it's plausibility score (inverse of the total loss). The details of this algorithm are given in the supplementary material.

\section{Experiments}

In this section we present two sets of experiments evaluating the proposed model. The first set of experiments (section~\ref{sec:intphys}) is on the IntPhys benchmark that is becoming the de facto standard for evaluating models of intuitive physics\footnote{www.intphys.com}~\cite{riochet_intphys:_2018}, and is currently used as evaluation in the DARPA Machine Common Sense program. The second set  of experiments (section~\ref{sec:future_prediction}) evaluates the accuracy of the predicted object trajectories and is inspired by the evaluation set-up used in~\cite{battaglia_interaction_2016} but here done in 3D with inter-object occlusions. 

\subsection{Evaluation on the IntPhys benchmark}
\label{sec:intphys}

\paragraph{Dataset.}
The Intphys Benchmark consists in a set of video clips in a virtual environment. Half of the videos depict possible event and half impossible.
They are organized in three blocks, each one testing for the ability of artificial systems to discriminate a class of physically impossible events. Block \texttt{O1} contains videos where objects may disappear with no reason, thus violating object permanence. In Block \texttt{O2}, objects' shape may change during the video, again without any apparent physical reason. In Block \texttt{O3}, objects may "jump" from one place to another, thus violating continuity of trajectories. Systems have to provide a plausibility score for each of the 12960 clips and are evaluated in terms of how well they can classify possible and impossible movies. Half of the impossible events ($6480$ videos) occur in plain sight, and are relatively easy to detect. The other half occurs under complete occlusion, leading to poor performance of current methods \cite{riochet_intphys:_2018,Smith2019ModelingEV}.

Along with the test videos, the benchmark contains an additional training set with 15000 videos, with various types of scenes, object movements and textures.
Importantly, the training set only consists in possible videos. Solving this task therefore cannot be done by learning a classifier or plausibility score from the training set.

\paragraph{System training.}
We use the training set to train the Compositional Rendering Network and a MaskRCNN object detector/segmenter from groundtruth object positions and segmentations. We also train the Recurrent Interaction Network to predict trajectories of object 8 frames in the future, given object positions in pairs of input frames. Once trained, we apply the scene graph proposal and optimization algorithm described above and derive the plausibility score which we take as the inverse of a plausibility loss.

\paragraph{Results.}
Table \ref{tab:intphys} reports error rates (smaller is better) for the three above mentioned blocks each in the visible and occluded set-up, with ``Total" reporting the overall error. We compare performance of our method with two strong baselines \citet{riochet_intphys:_2018} and the current state-of-the-art on Block \texttt{O1} \cite{Smith2019ModelingEV}. We observe a clear improvement over the two other methods, mainly explained by better predictions when impossible events are occluded (see \textit{Occluded} columns). In particular, results in the \textit{Visible} case are rather similar to \citet{riochet_intphys:_2018}, with a slight improvement of $2\%$ on \texttt{O1} and $6\%$ on \texttt{O3}. On the other hand, improvements on the \textit{Occluded} reach $33\%$ on \texttt{O1} and $21\%$ on \texttt{O2} clearly demonstrating our model can better deal with occlusions. We could not obtain the Visible/Occluded split score of \cite{Smith2019ModelingEV} by the time of the submission, thus indicating question marks in the Table \ref{tab:intphys}. On \texttt{O3/Occluded}, we observe that our model still struggles to detect correctly impossible events. Interestingly, the same pattern can be observed in human evaluation detailed in~\citet{riochet_intphys:_2018}, with a similar error rate in the Mechanical Turk experiment. This tends to show that detecting object "teleportation" under significant occlusions is more complex than other tasks in the benchmark. It would be interesting to confirm this pattern with other methods and/or video stimuli.
Overall results demonstrate a clear improvement of our method on the IntPhys benchmark, confirming its ability to follow objects and predict motion under long occlusions.

\begin{table*}[h]
\centering
\begin{small}
\begin{tabular}{lcccccccccccc}
&& \multicolumn{3}{c}{Block O1} && \multicolumn{3}{c}{Block O2} && \multicolumn{3}{c}{Block O3} \\
\toprule
            && Visible & Occluded & Total && Visible & Occluded & Total  && Visible & Occluded & Total  \\ 
\midrule
Ours   && \textbf{0.05}  & \textbf{0.19}   & \textbf{0.12}  && \textbf{0.11}  & \textbf{0.31} & \textbf{0.21} && \textbf{0.26} & \textbf{0.47} & \textbf{0.37} \\
\cite{riochet_intphys:_2018}   && 0.07  & 0.52   &  0.29  && \textbf{0.11}  & 0.52 &   0.31 &&  0.32  & 0.51 &   0.41 \\
\cite{Smith2019ModelingEV}   &&  -  & -  & 0.27  && -  & - &  -  && - & - & -  \\
\toprule
Human judgement  &&  0.18  & 0.30  & 0.24  && 0.22  & 0.29 &  0.25  && 0.28 & 0.47 & 0.37  \\
\bottomrule
\end{tabular}
\end{small}
\vspace*{-2mm}
\caption{{\bf Results on the IntPhys benchmark.} Relative classification error of our model compared to \cite{riochet_intphys:_2018} and \cite{Smith2019ModelingEV}, demonstrating large benefits of our method in scenes with significant occlusions ("Occluded"). Human judgement reports average errors of human judgements, as presented in \cite{riochet_intphys:_2018}.  Lower is better.}
\label{tab:intphys}
\end{table*}

\subsection{Evaluation on Future Prediction}
\label{sec:future_prediction}

In this section we investigate in more detail the ability of our model to learn to predict future trajectories of objects despite large amounts of inter-object occlusions. We first describe the dataset and experimental set-up, then discuss the results of object trajectory prediction under varying levels occlusion. Next, we report ablation studies comparing our model with several strong baselines. Finally, we report an experiment demonstrating that our model generalizes to real scenes.

\paragraph{Dataset.}

We use pybullet\footnote{\url{https://pypi.org/project/pybullet}} physics simulator to generate videos of a variable number of balls of different colors and sizes bouncing in a 3D scene (a large box with solid walls) containing a variable number of smaller static 3D boxes.
We generate five datasets, where we vary the camera tilt and the presence of occluders. 
In the first dataset (``Top view") we record videos with a top camera view (or \ang{90}), where the borders of the frame coincide with the walls of the box. In the second dataset (``Top view+occ"), we add a large moving object occluding 25\% of the scene. Finally, we decrease the camera viewing angle to \ang{45}, \ang{25} and \ang{15} degrees, which results in an increasing amount of inter-object object occlusions due to perspective projection of the 3D scene onto a 2D image plane.
Contrary to the previous experiment on IntPhys benchmark, we use the ground truth instance masks as the input to our model to remove potential effects due to errors in object detection. Additional details of the datasets and visualizations are given in the supplementary material.

\paragraph{Trajectory prediction in presence of occlusions.}
In this experiment we initialize the network with the first two frames. 
We then run a roll-out for $N$ consecutive frames using our model.
We consider prediction horizons of 5 and 10 frames, and evaluate the position error as a L2 distance between the predicted and ground truth object positions. L2 distance is computed in the 3D Cartesian scene coordinates so that results are comparable across the different camera tilts. 
Results are shown in Table~\ref{tab:poserror}.
We first note that our model (e. RecIntNet) significantly outperforms the linear baseline (a.), which is computed as an extrapolation of the position of objects based on their initial velocities. Moreover, the results of our method are relatively stable across the different challenging setups with occlusions by external objects (Top view+occ) or frequent self-occlusions in tilted views (tilt). This demonstrates the potential ability of our method to be trained from real videos where occlusions usually prevent reliable recovery of object states.

\begin{table*}
\small
\begin{center}
\begin{tabular}{lrrrrr}
\hline
                            & Top view & Top view+occ. & \ang{45} tilt & \ang{25} tilt & \ang{15} tilt \\
                            \hline
a. Linear baseline     &   47.6 / 106.0  &   47.6 /106.0 &   47.6 / 106.0 &   47.6 / 106.0 & 47.6 / 106.0 \\
b. MLP baseline        &   13.1 / 15.7  &   17.3 / 19.2  &   18.1 / 23.8  &   17.6 / 24.6  &  19.4 / 26.2\\
c. NoDyn-RecIntNet          &   21.2 / 46.2  &   23.7 / 46.7 &  22.5 / 42.8  &  23.1 / 43.3 &  24.9 / 44.4 \\
d. NoProba-RecIntNet                &   6.3 / 11.5 &   12.4 / 14.7 &   \textbf{8.0} / 15.9 &   8.12 / 16.3 &   11.2 / 19.6\\
e. RecIntNet (Ours)           &   6.3 / \textbf{9.2} &   \textbf{11.7} / \textbf{13.5} &   8.01 / \textbf{14.5} &   \textbf{8.1} / \textbf{15.0} &   11.2 / \textbf{18.1}\\
\hline
\end{tabular}
\end{center}
\vspace*{-2mm}
\caption{{\bf Object trajectory prediction in the synthetic dataset.} Average Euclidean L2 distance in pixels between predicted and ground truth positions, for a prediction horizon of 5 / 10 frames (lower is better). To compute the distance, the pixel-based x-y-d coordinates of objects are projected back in an untilted 200x200x200 reference Cartesian coordinate system.}
\label{tab:poserror}
\end{table*}

\paragraph{Ablation Studies.}
As an ablation study we replace the Recurrent Interaction Network ($RecIntNet$) in our model with a multi-layer perceptron (b. MLP baseline in Table~\ref{tab:poserror}). This MLP contains four hidden layers of size $180$ and is trained the same way as $RecIntNet$, modeling acceleration as described in equation \ref{eq:taylor}. To deal with the varying number of objects in the dataset, we pad the inputs with zeros. Comparing the MLP baseline (a.) with our model (e. RecIntNet) we observe that our $RecIntNet$ allows more robust predictions through time. 

As a second ablation study, we train the Recurrent Interaction Network without modeling acceleration (eq. \ref{eq:taylor}). This is similar to the model described in \cite{janner_reasoning_2018}, where object representation is not decomposed into position / velocity / intrinsic properties, but is rather a (unstructured) 256-dimensional vector. Results are reported in table~\ref{tab:poserror} (c. NoDyn-RecIntNet). Compared to our full approach (e.), we observe a significant loss in performance, confirming that modeling position and velocity explicitly, and having a constant velocity prior on motion (given by \ref{eq:taylor}) improves future predictions.

As a third ablation study, we train a deterministic variant of $RecIntNet$, where only the sequence of states is predicted, without the uncertainty term $\tau$ (please see more details in the Supplementary). The loss considered is the mean squared error between the predicted and the observation state.
Results are reported in table~\ref{tab:poserror} (d. NoProba-RecIntNet). The results are slightly worse than our model handling uncertainty (d. NoProba-RecIntNet), but close enough to say that this is not a key feature for modeling 5 or 10 frames in the future. In qualitative experiments, however, we observed more robust long-term predictions with uncertainty in our model.

\begin{figure}[h]
\begin{center}
\includegraphics[width=1\linewidth]{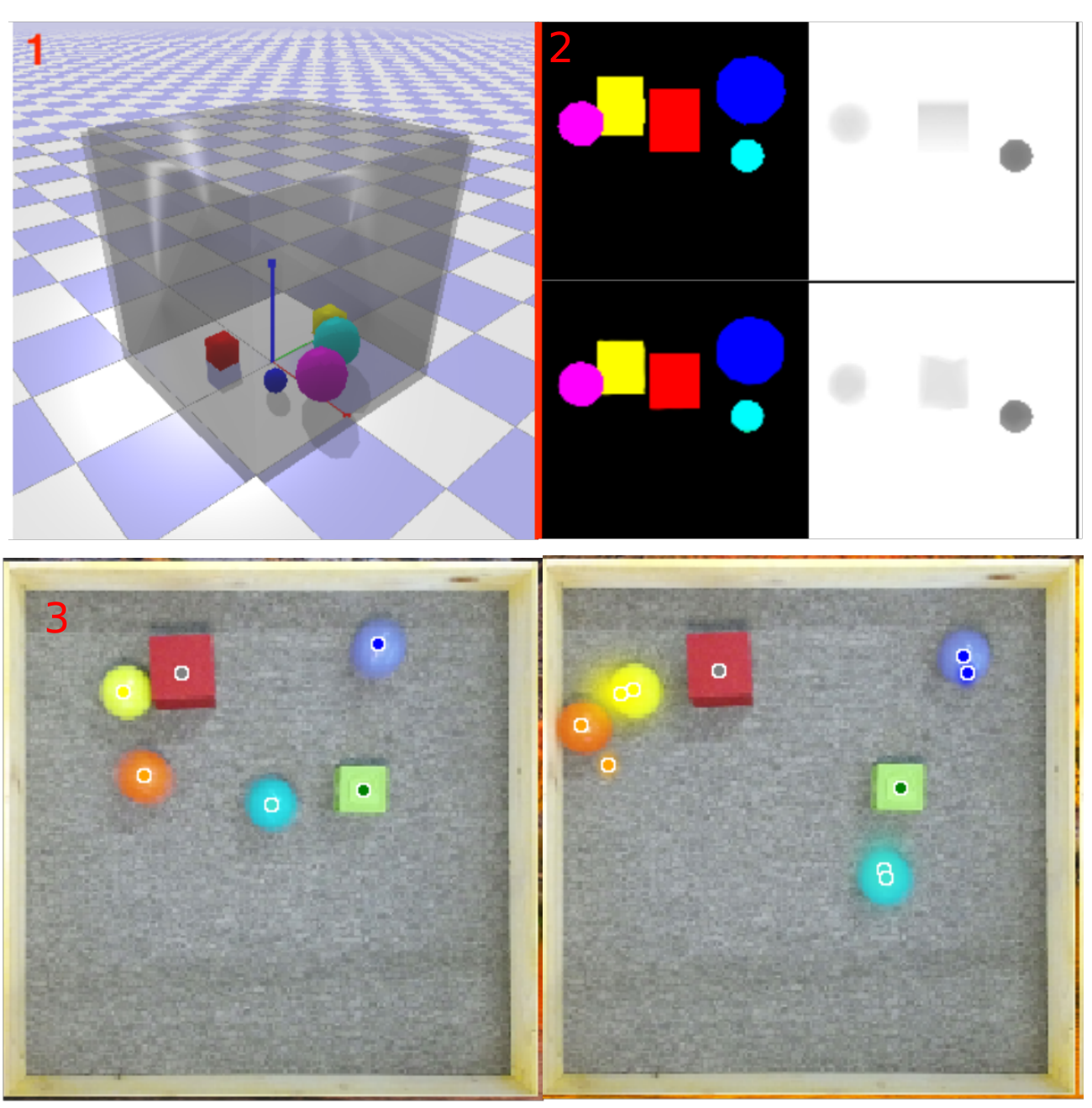}
\end{center}
\caption{{\bf Images from the Future Prediction experiment} 
1: An overview of the pybullet scene. 2: Sample video frames (instance mask + depth field) from our datasets (top) together with predictions obtained by our model (bottom), taken from from the tilted \ang{25} experiments. 3: example of prediction for a real video, with a prediction span of 8 frames. The small colored dots show the predicted positions of objects together with the estimated uncertainty shown by the colored “cloud”. The same colored dot is also shown in the (ground truth) center of each object. The prediction is correct when the two dots coincide. (see  
\href{https://drive.google.com/open?id=1Qc8flIAxUGzfRfeFyyUEGXe6J5AUGUjE}{additional videos}).}
\label{fig:qualitativeresults}
\end{figure}

\paragraph{Generalization to real scenes.}
We test the model trained on top-view synthetic Pybullet videos (without finetuning the weights) on a dataset of 22 real videos containing a variable number of colored balls and blocks in motion recorder with a Microsoft Kinect2 device. Example frames from the data are shown in figure \ref{fig:qualitativeresults}. Results are reported in the supplementary and demonstrate that our model generalizes to real data and show clear improvements over the linear and MLP baselines.

\paragraph{Additional results in the supplementary material.}
In addition to the forward prediction, we evaluate our method on the task of following objects in the scene. Details and results can be found in the supplementary material (section 5).

\section{Discussion}
Learning the physics of simple macroscopic object dynamics and interactions is a relatively easy task when ground truth coordinates are provided to the system, and techniques like Interaction Networks trained with a future frame prediction loss are quite successful~\cite{battaglia_interaction_2016,mrowca_flexible_2018}. In real-life applications, the physical state of objects is not available and has to be inferred from sensors. In such case inter-object occlusions make these observations noisy and sometimes missing. 

Here we present a probabilistic formulation of the intuitive physics problem, where observations are noisy and the goal is to infer the most likely underlying object states. This physical state is the solution of an optimization problem involving i) a physics loss: objects states should be coherent in time, and ii) a render loss: the resulting scene at a given time should match with the observed frame. We present a method to find an approximate solution to this problem, that is compositional (does not restrict the number of objects) and handles occlusions. We show its ability to learn object dynamics and demonstrate it outperforms existing methods on the intuitive physics benchmark IntPhys.

A second set of experiments studies the impact of occlusions on intuitive physics learning. During training, occlusions act like missing data because the object position is not available to the model. However, we found that it is possible to learn good models compared to baselines, even in challenging scenes with significant inter-object occlusions.  We also notice that projective geometry is not, in and of itself, a difficulty in the process. Indeed, when an our dynamics model is fed, not with 3D Cartesian object coordinates, but with a 2.5D projective referential such as the xy position of objects in a retina (plus depth), the accuracy of the prediction remains unchanged compared with the Cartesian ground truth. Outcomes of these experiments can be seen in the google drive (\href{https://drive.google.com/open?id=1Qc8flIAxUGzfRfeFyyUEGXe6J5AUGUjE}{link}).  This work, along with recent improvement of object segmentation models \cite{he_mask_2018} put a first step towards learning intuitive physics from real videos.

Further work needs to be done to fully train this system end-to-end, in particular, by learning the renderer and the interaction network jointly. This could be done within our probabilistic framework by improving the initialization step of our system (scene graph proposal). Instead of using a relatively simple heuristics yielding a single proposal per video clip, one could generate multiple proposals (a decoding lattice) that would be reranked with the plausibility loss. This would enable more robust joint learning by marginalizing over alternative event graphs instead of using a single point estimate as we do here. Finally object segmentation itself could be learned jointly, as this would allow exploiting physical regularities of the visual world as a bootstrap to learn better visual representations.

\section{Acknowledgements}

We would like to thank Malo Huard for his help in implementing the renderer, and anonymous reviewers for their helpful comments.

This work was partly supported by the European Regional Development Fund under the project IMPACT (reg. no. CZ.02.1.01/0.0/0.0/15\_003/0000468) and by the French government under management of Agence Nationale de la Recherche as part of the "Investissements d'avenir" program, reference ANR-19-P3IA-0001 (PRAIRIE 3IA Institute).

\bibliography{zotero,more}
\bibliographystyle{icml2020}

\title{Supplementary material}

\author{}
\date{}

\maketitle

This supplementary material: (i) describes the provided supplementary videos (section~\ref{sec:videos}), (ii) provides additional training details (section~\ref{sec:training}), (iii) explains in more depth the \textit{event decoding} procedure defined in section 3.4 in the main paper (section~\ref{sec:decoding}) (iv) gives details of the datasets used in the subsection 4.2 (section~\ref{sec:datasets}), (v) provides additional ablation studies and comparisons with baselines (sections~\ref{ablation},~\ref{sec:rollout},~\ref{sec:real},~\ref{sec:pixelerror}). 

\section{ Description of supplementary videos}
\label{sec:videos}

In this section we present qualitative results of our method on different datasets. We first show videos from IntPhys benchmark, where inferred object states are depicted onto observed frames. Then we show differents outputs on the pybullet datasets, for different levels of occlusions. Finally we present examples of predictions from our Recurrent Interaction Network on real scenes.

The videos are in the google drive: \url{https://drive.google.com/open?id=1Qc8flIAxUGzfRfeFyyUEGXe6J5AUGUjE} in the \texttt{videos/} subdirectory. Please see also the \texttt{README} slideshow in the same directory.

\subsection{IntPhys benchmark}

The Intphys Benchmark consists in a set of video clips in a virtual environment. Half of the videos are possible event and half are impossible, the goal being to discriminate the two.

In the following we show impossible events, along with outputs of our event decoding method. Our dynamics and rendering models predict future frames (masks) in the videos, which are compared with the observed masks (pre-trained detector). This allows us to derive a plausibility loss used to discriminate possible and impossible events (see section 4.1).

\begin{itemize}
 \item \textbf{occluded\_impossible\_*.mp4} show examples of impossible videos from the IntPhys benchmark, along with visualization of our method. Each video contains four splits; on \underline{top/left} is shown the raw input frame; on \underline{bottom/left} is the mask obtained from the raw frame with the pre-trained mask detector (which we call \textit{observed mask}); on \underline{top/right} is the raw frame with superimposed output physical states predicted by our method; on \underline{bottom/right} is the reconstructed mask obtained with the Compositional Renderer (which we call \textit{predicted mask}). Throughout the sequence, our method predicts the full trajectory of objects. When an object should be visible (i.e. not behind an occluder), the renderer predicts correctly its mask. If at the same time the object has disappeared from the observed mask, or changed too much in position or shape, it causes a mismatch between the predicted and the observed masks, hence a higher plausibility loss. This plausibility loss is use for the classification task of IntPhys benchmark (see quantitative results in main paper, section 4.1).
 \item \textbf{visible\_impossible\_*.mp4} show similar videos but with impossible events occurring in the "visible" (easier) task of the IntPhys benchmark. 
 \item \textbf{intphys\_*.mp4} show object following in the IntPhys training set.
 \end{itemize}
 
\subsection{Pybullet experiments}
 
 We present qualitative results on our Pybullet dataset. We construct videos including a various number of objects with different points of view and increasing levels of camera tilts introducing inter-object occlusions. First, we show predicted physical states drawn on object states, to demonstrate the ability of the method to track objects under occlusions. Then we show videos of long rollouts where, from one pair of input frames, we predict a full trajectory and render masks with the Compositional Neural Renderer.

 \begin{itemize}
 \item \textbf{scene\_overview.mp4} shows raw videos of the entire environment.
 \item \textbf{tracking\_occlusions\_*.mp4} show examples of position prediction through complete occlusions, using our event decoding procedure. This shows that our model can keep track of the object identity through complete occlusions, mimicking ``object permanence".
 \item \textbf{one\_class*.mp4} show different examples of our model following motion of multiple objects in the scene. All balls have the same color which makes them difficult to follow in case of mutual interactions. Videos come from tilted \ang{25} experiments, which are the most challenging because they include inter-object occlusions. Dots represent the predicted position of each object, the color being its identity. Our model shows very good predictions with small colored markers (dots) well centered in the middle of each object, with marker color remaining constant for each object preserving the object identity during occlusions and collisions. \textbf{one\_class\_raw*.mp4} show rendered original views of the same dynamic scenes but imaged from a different viewpoint for better understanding.
 \item \textbf{rollout\_0.mp4}, \textbf{rollout\_1.mp4} show three different prediction roll-outs of the Recurrent Interaction Network (without event decoding procedure). From left to right: ground truth trajectories, our model trained of state, our model trained on masks, our model trained on masks with occlusions during training. Rollout length is 20 frames.
 \item \textbf{rollout\_tilt*\_model.mp4} and \textbf{rollout\_tilt*\_groundtruth.mp4} show the same dynamic scene but observed with various camera tilts (e.g. \textbf{tilt45\_model.mp4} show a video for a camera tilt of 45 degrees). \textbf{*\_model.mp4} are predicted roll-outs of our Recurrent Interaction Network ($RecIntNet$), without event decoding. \textbf{*\_groundtruth.mp4} are the corresponding ground-truth trajectories, rendered with the \textit{Compositional Rendering Network}.
 \item \textbf{rollout\_pybullet\_*.mp4} show free roll-outs (no event decoding) on synthetic dataset.
 \end{itemize}
 
 \subsection{Real videos}
 
 \begin{itemize}
 \item \textbf{rollout\_real\_*.mp4} show generalization to real scenes.
\end{itemize}

\section{ Training details}
\label{sec:training}

This section gives details of the offline pre-training of the Compositional Rendering Network and detailed outline of the algorithm for training the Recurrent Interaction Network.

\paragraph{Pre-Training the Compositional Rendering Network.} We train the neural renderer to predict mask and depth $\hat{M}_t,\hat{D}_t$ from a list of objects $[p_x,p_y,d,\bm{c}]$ where
$p_x,p_y$ are x-y coordinates of the object in the frame, $d$ is the distance between the object and the camera and $\bm{c}$ is a vector for intrinsic object properties containing the size of the object, its class (in our experiments a binary variable for whether the object is a ball, a square or an occluder) and its color as vector in $[0,1]^3$. In IntPhys benchmark, occluders are not modeled with a single point $[p_x,p_y,d,\bm{c}]$ but with four points $[p_x^k,p_y^k,d^k],k=1..4$ corresponding to the four corners of the quadrilateral. These four corners are computed from the occluder instance mask, after detecting contours and applying Ramer–Douglas–Peucker algorithm to approximate the shape with a quadrilateral.

The target mask is a $128\times128$ image where each pixel value indicates the index of the corresponding object mask (0 for the background, $i\in{1..N}$ for objects). The loss on the mask is negative log-likelihood, which corresponds to the average classification loss on each pixel
\begin{align}
\small
 L_{\text{mask}}(\hat{M}, M) = \sum_{i \leq h, j \leq w} \sum_{n \leq N} \bm{1}(M_{i,j}=n)log(\hat{M}_{i,j,n}),
 \label{eq:lossmask}
\end{align}
where the first sum is over individual pixels indexed by $i$ and $j$, the second sum is over the individual objects indexed by $n$, $\forall \hat{M} \in [0,1]^{h\times w\times N}$ are the predicted (soft-) object masks, and $\forall M \in [\![1,N[\!]^{h \times w}$ is the scene ground truth mask containing all objects. 

The target depth map is a $128\times128$ image with values being normalized to the [-1,1] interval during training. The loss on the depth map prediction is the mean squared error
\begin{align}
 L_{\text{depth}}(\hat{D}, D) = \sum_{i \leq h, j \leq w} (\hat{D}_{i,j} - D_{i,j})^2, 
 \label{eq:lossdepth}
\end{align}
where $\forall \hat{D}$ and $D \in \mathbb{R}^{h\times w}$ are the predicted and ground truth depth maps, respectively. 
The final loss used to train the renderer is the weighted sum of losses on masks and depth maps, $L = 0.7 * L_{mask} + 0.3 * L_{depth}$. We use the \texttt{Adam} optimizer with default parameters, and reduce learning rate by a factor $10$ each time the loss on the validation set does not decrease during $10$ epochs. We pre-train the network on a separate set of 15000 images generated with \texttt{pybullet} and containing similar objects as in our videos.

\paragraph{Training details of the Recurrent Interaction Network.}

From a sequence of L frames with their instance masks we compute objects position, size and shape (see section 3.2 in the main body). Initial velocities of objects are estimated as the position deltas between the first two positions. This initial state (position, velocity, size and shape of all objects) is given as input of the Recurrent Interaction Network to predict the next L-2 states. The predicted L-2 positions are compared with observed object positions. The sum of prediction errors (section 3.3 in core paper) is used as loss to train parameters of the Recurrent Interaction Network. Optimization is done via gradient descent, using Adam with learning rate $1e-3$, reducing learning by a factor of 10 each time loss on validation plateaus during 10 epochs. We tried several sequence lengths ($4$, $6$, $10$), $10$ giving the most stable results. During such sequence, when an object was occluded (thus position not being observed), we set its loss to zero. 

\section{Event Decoding}
\label{sec:decoding}

\begin{figure*}[h!]
\begin{center}
\includegraphics[width=1\linewidth]{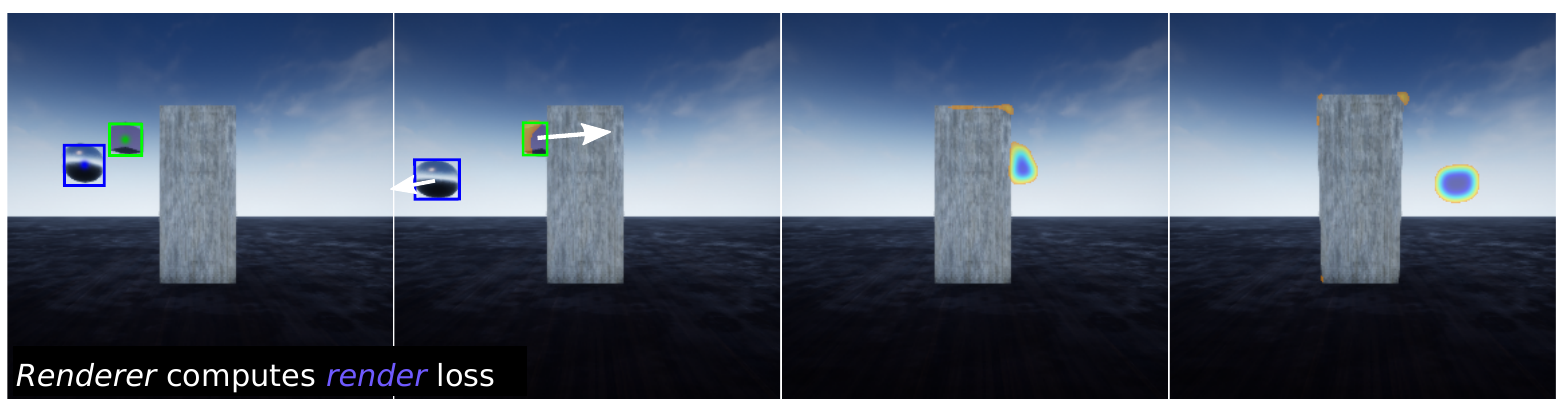}
\caption{{\bf Video example from the IntPhys benchmark.} Four frames from a video in block \texttt{O1}, with superimposed heatmaps. Heatmaps (colored blobs) correspond to the difference, per pixel, between the predicted and the observed object mask. In these video, a cube moves from left to right  but disappears behind the occluder. The Recurrent Interaction Network predicts correctly its motion behind the occluder and the Compositional Renderer reconstructs its mask. The fact that the object is absent in the observed mask leads to a large \textit{render} loss, illustrated by the high heatmap values (violet) at the position where the ball is expected to be. }
\label{fig:decoding_1}
\end{center}
\end{figure*}

\begin{figure*}[h!]
\begin{center}
\includegraphics[width=0.75\linewidth]{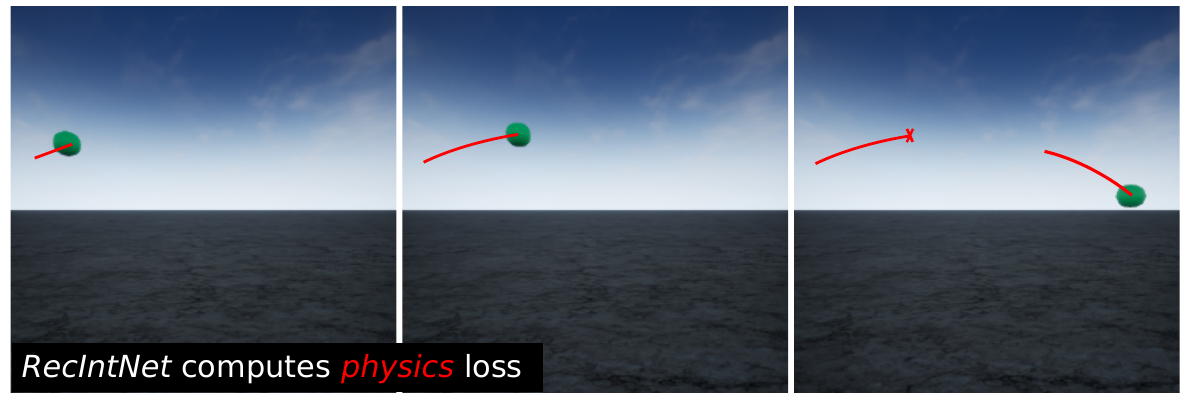}
\caption{{\bf Video example from the IntPhys benchmark.} Three frames from a video in block \texttt{O2}, where an object "jumps" from one place to another. The graph proposal phase returns the right trajectory of the object but the Recurrent Interaction Network returns a high \textit{physics} loss at the moment of the jump, because the observed position is far from the predicted one.}
\label{fig:decoding_2}
\end{center}
\end{figure*}

The detailed outline of the event decoding procedure described in section \ref{sec:decoding} of the main paper is given in Algorithm~\ref{alg:procedure}.  Two example figures (Figure \ref{fig:decoding_1} \& \ref{fig:decoding_2}) gives an intuition behind the \textit{render} and \textit{physics} losses.

\begin{algorithm}[tbp]
\small
\SetAlgoLined
\KwData{\\
\Indp $T$: length of the video\;\\
$f_t,m_t~t=1..T$: videos frames, segmentation masks\;\\
\texttt{Detection($m_t$)}: returns centroid and size of instance masks\;\\
\texttt{RecIntNet}: Pre-trained Recurrent Interaction Network\;\\
\texttt{Rend}: Pre-trained Neural Renderer\;\\
\texttt{ClosestMatch(a,b)}: for a, b two lists of objects, computes the optimal ordering of elements in b to match those in a \;\\
$0 < \lambda < 1$: weighting physical and visual losses\;\\
}
\KwResult{\\
\Indp Estimated states $s_{1...T}$\;\\
Plausibility loss for the video\;\\
}
\vspace{0.7cm}
\textbf{Initialization}:\\
 $d_{t=1..T} = \text{Detection}(f_t)$\;
 $n_t \leftarrow (\#\{d_t\}, \text{mean}_t\textit{size}(d_t)$) \;\\
 $t^* \leftarrow \argmax_{t}({n_t + n_{t+1}})$\;\\
 \CommentSty{//$(t^*,t^*+1)$ is the pair of frames containing the maximum number of objects (with the max number of visible pixels in case of equality).}\\
 ($p_{t^*},p_{t^*+1}) \leftarrow (d_{t^*}, \text{ClosestMatch}(d_{t^*}, d_{t^*+1}))$\;\\
 \CommentSty{//Rearange $d_{t^*+1}$ to have same object ordering as in $d_{t^*}$}.\\

\textbf{Graph Proposal}:\\
\CommentSty{//$t^*$ is a good starting point for parsing the scene (because we observe most of the objects during two consecutive frames). We use RecIntNet to predict the next position of each object, which we link to an object detection. Repeating this step until the end of the video returns object trajectory candidates.}\\
$v_{t^*+1} \leftarrow p_{t^*+1} - p_{t^*}$\;\\
$s_{t^*+1} \leftarrow [p_{t^*+1},v_{t^*+1}]$\;\\
\For{$t \in \{t^*+1,..,T\}$}{ 
 $\hat{s}_{t+1} \leftarrow \text{RecIntNet}(s_t)$\;\\
 $s_{t+1} \leftarrow \text{ClosestMatch}(\hat{s}_{t+1}, d_{t+1})$\;\\
}
\CommentSty{//Backward: do the same from $t^*$ to 1.}
\\
\textbf{Differentiable optimization}:\\
 \CommentSty{//$\hat{s}_{t=1..T}$ is a sequence of physical states. At every time step $t$ it contains the position, velocity, size and shape of all objects, in the same order. Due to occlusions and detection errors, it is sub-optimal and can be refined to minimize equation 3 in the main paper.}\\
$\text{Loss}_\text{physics}(s) \leftarrow \sum_{t=1}^T \|\hat{s}_{t+1} - s_{t+1}\|^2$\;\\
$\text{Loss}_\text{visual}(s) \leftarrow \sum_{t=1}^T \texttt{NLL}(\texttt{Rend}(s_t), m_t)$\;\\
$\text{Loss}_\text{plausibility}(s) \leftarrow \lambda\text{Loss}_\text{physics}(s) + (1-\lambda)\text{Loss}_\text{visual}(s)$\;
 $(\text{Estimated states}, \text{plausibility loss}) \leftarrow \texttt{SGD}_s(\text{Loss}_\text{plausibility}(s))$\;
 \\\CommentSty{//with $lr=1e-3$ and $n_\text{steps}=1000$}\;

\caption{{\bf Event decoding procedure}}
\label{alg:procedure}
\end{algorithm}

\section{ Datasets}
\label{sec:datasets}

\begin{figure*}[h!]
\begin{center}
\includegraphics[width=1\linewidth]{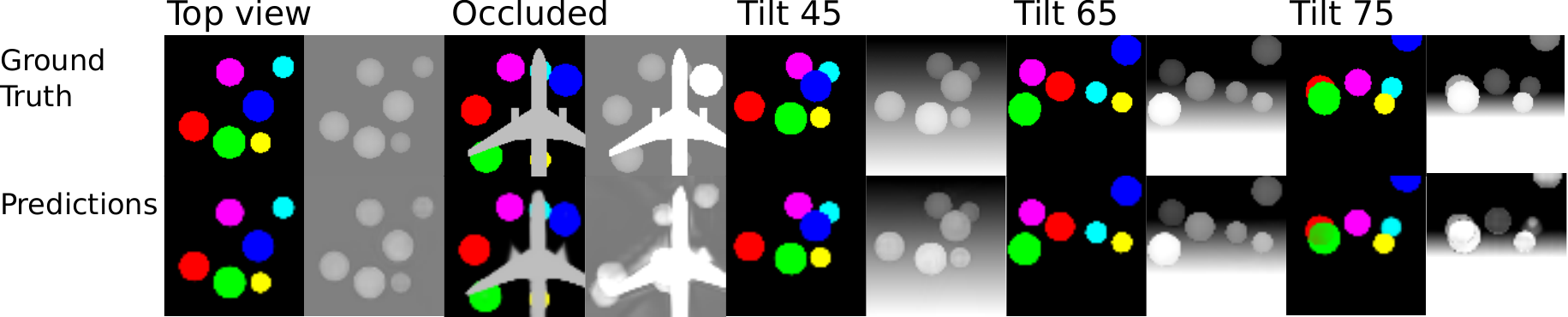}
\end{center}
\caption{ Sample video frames (instance mask + depth field) from our datasets (top) together with predictions obtained by our model (bottom). Taken from the top-view, occluded and tilted experiments. {\bf Please see additional video results in the google drive \url{https://drive.google.com/open?id=1Qc8flIAxUGzfRfeFyyUEGXe6J5AUGUjE}.}}
\end{figure*}

To validate our model, we use pybullet\footnote{\url{https://pypi.org/project/pybullet}} physics simulator to generate videos of variable number of balls of different colors and sizes bouncing in a 3D scene (a large box with solid walls) containing a variable number of smaller static 3D boxes. We generate five dataset versions, where we vary the camera tilt and the presence of occluders. All experiments are made with datasets of 12,000 videos of 30 frames (with a frame rate of 20 frames per second). For each dataset, we keep 2,000 videos separate to pre-train the renderer, $9,000$ videos to train the physics predictor and $1,000$ videos for evaluation. Our scene contains a variable number of balls (up to 6) with random initial positions and velocities, bouncing against each other and the walls. Initial positions are sampled from a uniform distribution in the box $[1,200]^2$, all balls lying on the ground. Initial velocities along $x$ and $y$ axes are sampled in $\textit{Unif}([-25,25])$ units per frame, initial velocity along $z$-axis is set to $0$. The radius of each ball is sampled uniformly in $[10,40]$. Scenes also contain a variable number of boxes (up to 2) fixed to the floor, against which balls can collide. Contrary to \cite{battaglia_interaction_2016} where authors set a frame rate of 1000 frames per second, we sample 30 frames per second, which is more reasonable when working with masks (because of the computation cost of mask prediction).

\paragraph{Top-view.}
In the first dataset we record videos with a top camera view, where the borders of the frame coincide with the walls of the box. Here, initial motion is orthogonal to the camera, which makes this dataset very similar to the 2D bouncing balls datasets presented in \cite{battaglia_interaction_2016} and \cite{watters_visual_2017}. However, our dataset is 3D and because of collisions and the fact that the balls have different sizes, balls can jump on top of each other, making occlusions possible, even if not frequent. 

\paragraph{Top-view with Occlusions.}
To test the ability of our method to learn object dynamics in environments where occlusions occur frequently, we record the second dataset including frequent occlusions. We add an occluder to the scene, which is an object of irregular shape (an airplane), occluding $25\%$ of the frame and moving in 3D between the balls and the camera. This occluder has a rectilinear motion and goes from the bottom to the top of the frame during the whole video sequence. Sample frames and rendered predictions can be found in the supplementary material.

\paragraph{Tilted-views.}
In three additional datasets we keep the same objects and motions but tilt the camera with angles of \ang{45}, \ang{65} and \ang{75} degrees. Increasing the tilt of the camera results in more severe inter-object occlusions (both partial and complete) where the balls pass in front of each other, and in front and behind the static boxes, at different distances to the camera. In addition, the ball trajectories are becoming more complex due to increasing perspective effects.
In contrary to the top-view experiment, the motion is not orthogonal to the camera plane anymore, and depth becomes crucial to predict the future motion.

\section{Ablation studies}
\label{ablation}

For the purpose of comparison, we also evaluate three models trained using ground truth object states. Results are shown in table \label{ref:gtposerror}. Our Recurrent Interaction Network trained on ground truth object states gives similar results to the model of~\cite{battaglia_interaction_2016}. As expected, training on ground truth states (effectively ignoring occlusions and other effects) performs better than training from object masks and depth.
\begin{table}[h]
\footnotesize
\begin{center}
\begin{tabular}{lrrrr}
\hline
     & Top view & \ang{45} tilt & \ang{25} tilt & \ang{15} tilt \\   
%     & & occlusion & & & \\
     \hline
\multicolumn{5}{c}{}\\
\scriptsize{NoProba-RIN}      & 4.76 / 9.72 & 6.21 / 10.0 & 5.2 / 12.2 & 7.3 / 13.8 \\
RIN     & 4.5 / 9.0 & 6.0 / 9.6 & 5.2 / 12.2 & 7.3 / 13.2 \\
\citeyear{battaglia_interaction_2016}** & 3.6 / 10.1 & 4.5 / 9.9 & 4.5 / 11.0 & 5.3 / 12.3\\
\hline
\end{tabular}
\end{center}
\caption{Average Euclidean (L2) distance (in an untilted 200x200x200 reference Cartesian coordinate system) between predicted and ground truth positions, for a prediction horizon of 5 frames / 10 frames, trained on ground truth positions. **\cite{battaglia_interaction_2016} is trained with more supervision, since target values include ground truth velocities, not available to other methods.
}
\label{tab:gtposerror}
\end{table}

\section{Roll-out results}
\label{sec:rollout}

\begin{figure}
\centering
\begin{subfigure}{.45\textwidth}
 \centering
 \includegraphics[width=1\linewidth]{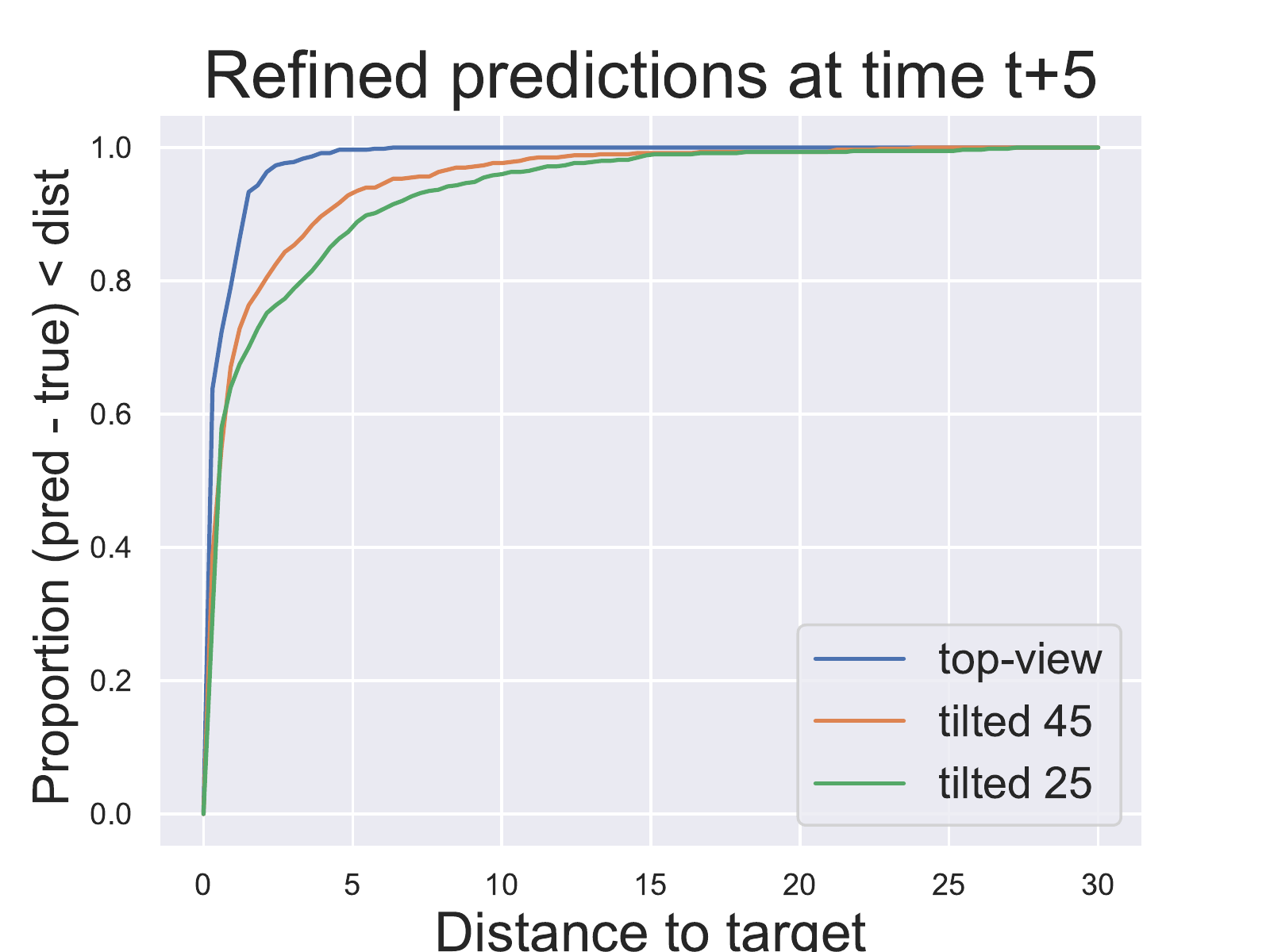}
\end{subfigure}
\begin{subfigure}{.45\textwidth}
 \centering
 \includegraphics[width=1\linewidth]{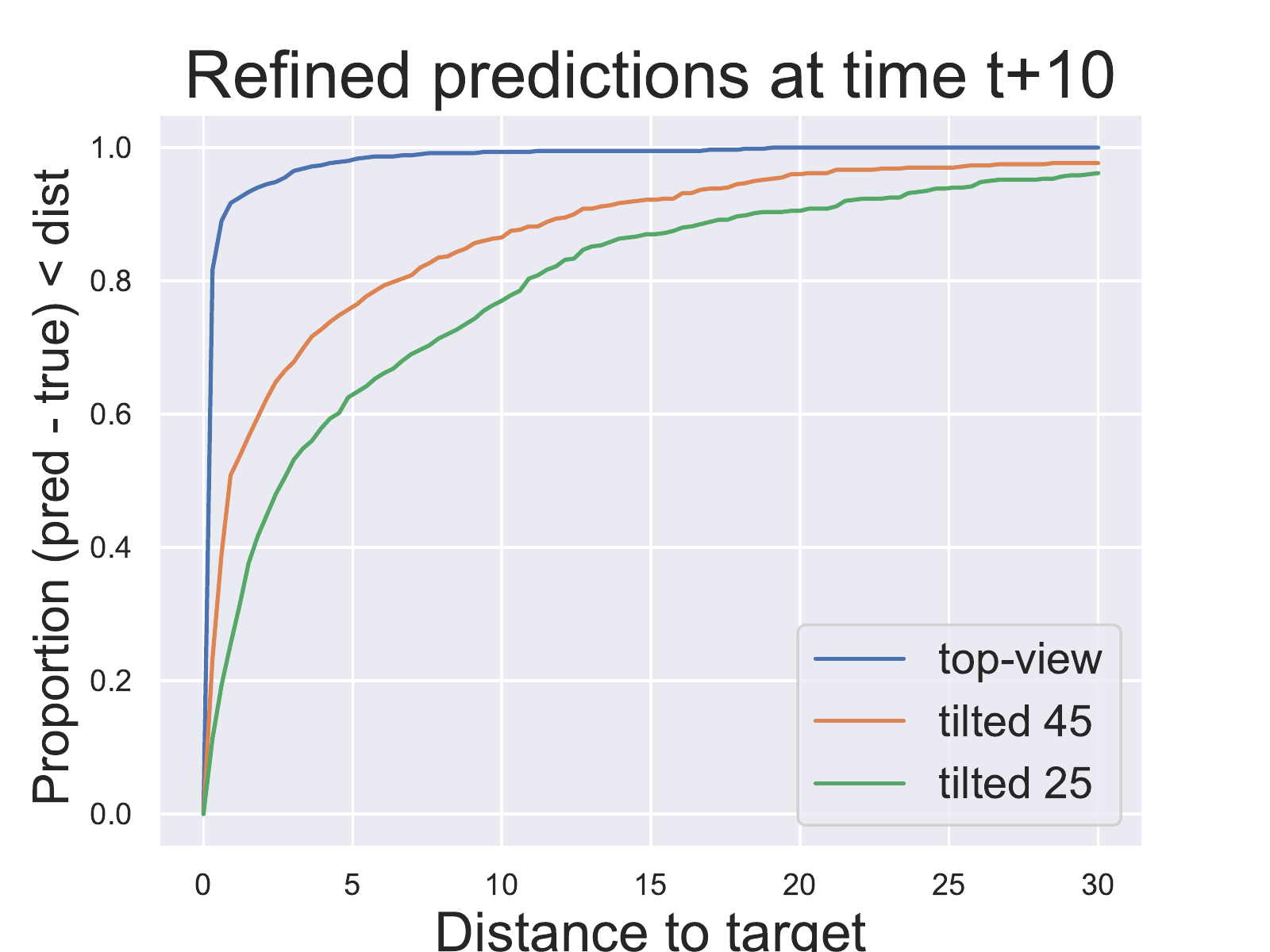}
\end{subfigure}
\begin{subfigure}{.45\textwidth}
 \centering
 \includegraphics[width=1\linewidth]{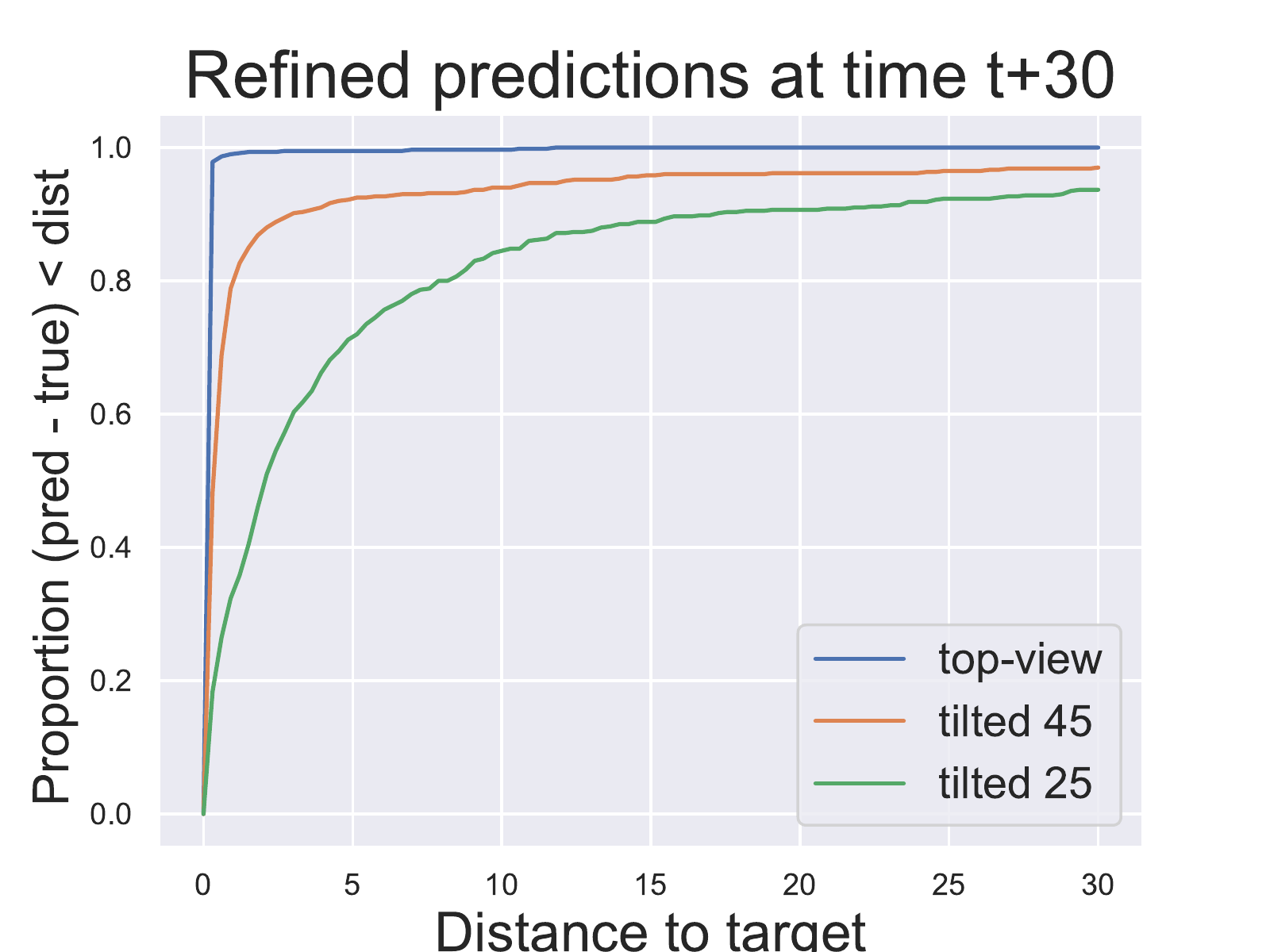}
\end{subfigure}
\caption{Proportion of correctly followed objects (y-axis) as a function of the distance error in pixels (x-axis) for our approach using online event decoding. The different plots correspond to rollout lengths of 5 (left), 10 (middle) and 30 (right) frames. Different curves correspond to different camera view angles (top-view, tilted 45 degrees and tilted 25 degrees). In this experiment, all objects have the same shape and color making the task of following the same object for a long period of time very challenging. The plots demonstrate the success of our method in this very challenging scenario with object collisions and inter-object occlusions. For example, within a distance threshold of 20 pixels, which corresponds to the size of the smallest objects in the environment, our approach correctly follows more than 90\% of objects during the rollout of 30 frames in all three considered camera viewpoints (top-view, 45 degrees and 25 degrees). {\bf Please see also the supplementary videos ``one\_class*.mp4".}
\label{fig:tracking}
}
\end{figure}

We evaluate our model on \textit{object following}, applying an online variant of the scene decoding procedure detailed in \ref{sec:decoding}. This online variant consists in applying the state optimization sequentially (as new frames arrive), instead of on the full sequence. For each new frame, the state prediction $\hat{s}_{t+1}$ given by $RecIntNet$ is used to predict a resulting mask. This mask is compared to the observation, and we apply directly the final step in Algorithm \ref{alg:procedure} (Differentiable optimization). It consists in minimizing  $\lambda\text{Loss}_\text{physics}(s) + (1-\lambda)\text{Loss}_\text{visual}(s)$ via gradient descent over the state $s$. During full occlusion, the position is solely determined by $RecIntNet$, since $\text{Loss}_\text{render}$ has a zero gradient. When the object is completely or partially visible, the $\text{Loss}_\text{render}$ in the minimization make the predicted state closer to its observed value. To test object following, we measure the accuracy of the position estimates across long sequences (up to 30 frames) containing occlusions. 
Table \ref{tab:tracking} shows the percentage of object predictions that diverge by more than an object diameter (20 pixels) using this method. The performance is very good, even for tilted views. In Figure \ref{fig:tracking}, we report the proportion of correctly followed objects for different rollout lengths (5, 10 and 30 frames) as a function of the distance error (pixels). Note that the size of the smallest object is around 20 pixels.

\begin{table}[h]
\begin{center}
\begin{tabular}{lccccc}
\hline
Synthetic videos & 5 fr. & 10 fr. & 30 fr. \\
\hline
 Ours, top view & 100 & 100 & 100  \\
 Ours, \ang{45} tilt & 99.3 & 96.2 & 96.2  \\
 Ours, \ang{25} tilt & 99.3 & 90.1 & 90.1  \\
 Linear motion baseline & 81.1 & 67.8 & 59.7 \\
\hline
\end{tabular}
\end{center}
\caption{Percentage of predictions within a 20-pixel neighborhood around the target as a function of rollout length measured by the number of frames. 20 pixels corresponds to the size of the smallest objects in the dataset.}
\label{tab:tracking}
\end{table}

\section{ Experiment with real videos}
\label{sec:real}

We construct a dataset of 22 real videos, containing a variable number of colored balls and blocks in motion. Videos are recorded with a Microsoft Kinect2 device, including RGB and depth frames. The setup is similar to the one generated with Pybullet, recorded with a top camera view and containing 4 balls and a variable number of static blocks (from 0 to 3). Here again, the borders of the frame coincide with the walls of the box.
Taking as input object segmentation of the first two frames, we use our model to predict object trajectories through the whole video (see Figure \ref{fig:qualitativeresults}). We use the model trained on top-view synthetic Pybullet videos, without fine-tuning weights. We measure the error between predictions and ground truth positions along the roll-out. Results are shown in Table \ref{tab:realposerror} and clearly demonstrate that out approach outperforms the linear and MLP baselines and makes reasonable predictions on real videos.

\begin{table}
\small
\begin{center}
\begin{tabular}{lccccc}
\toprule
Model && Linear & MLP & Proba-RecIntNet (ours)\\    
L2 dist. to target && 28/71 & 19/43 & \textbf{12/22} \\
\bottomrule
\end{tabular}
\end{center}
\vspace*{-4mm}
\caption{{\bf Trajectory prediction on real videos.} Average Euclidean (L2) distance (in pixels in a 200 by 200 image) between predicted and ground truth positions, for a prediction horizon of 5 frames / 10 frames.}
\label{tab:realposerror}
\end{table}

\begin{figure*}
\begin{center}
\includegraphics[width=1\linewidth]{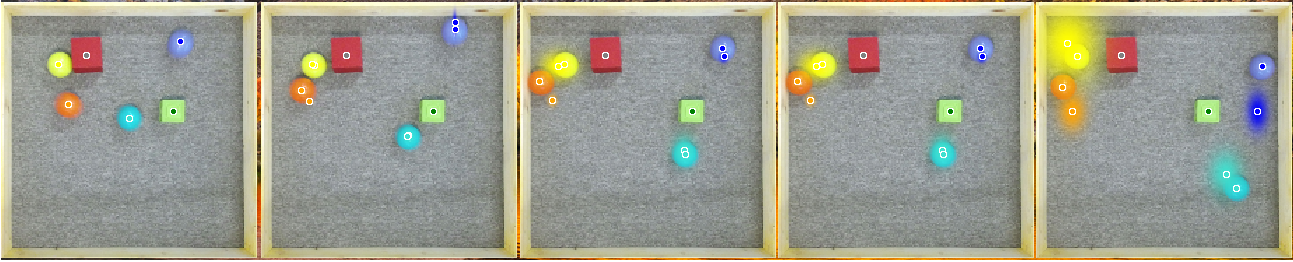}
\end{center}
\caption{Example of prediction for a real video, with a prediction span of 10 frames. The small colored dots show the predicted positions of objects together with the estimated uncertainty shown by the colored “cloud”. The same colored dot is also shown in the (ground truth) center of each object. The prediction is correct when the two dots coincide. (see \href{https://drive.google.com/open?id=1Qc8flIAxUGzfRfeFyyUEGXe6J5AUGUjE}{additional videos}).}
\label{fig:real}
\end{figure*}

\section{ Future prediction (pixels): Comparison with baselines}
\label{sec:pixelerror}

We evaluate the error of the mask and depth prediction, measured by the training error described in detail in \ref{sec:training}.
Here, we compare our model to a CNN autoencoder~\cite{riochet_intphys:_2018}, which directly predicts future masks from current ones, without explicitly modelling dynamics of the individual objects in the scene. Note this baseline is similar to~\cite{lerer_learning_2016}.
Results are shown in Table S1.
As before, the existence of external occluders or the presence of tilt degrades the performance, but even in this case, our model remains much better than the CNN autoencoder of~\cite{riochet_intphys:_2018}.

\begin{table*}
\small
\begin{center}
\begin{tabular}{lccccc}
\hline
     & Top view & Top view+ & \ang{45} tilt & \ang{25} tilt & \ang{15} tilt \\      
     & & occlusion & & & \\
\hline
CNN autoencoder \cite{riochet_intphys:_2018} & 0.0147 & 0.0451 & 0.0125 & 0.0124 & 0.0121 \\
NoProba-RIN & 0.0101 & 0.0342 & 0.0072 & 0.0070 & 0.0069 \\
Proba-RIN & 0.0100 & 0.0351 & 0.0069 & 0.0071 & 0.0065 \\
\hline
\end{tabular}
\end{center}
\caption*{Aggregate pixel reconstruction error for mask and depth, for a prediction span of two frames. This error is the loss used for training (described in the supplementary material). It is a weighted combination of mask error (per-pixel classification error) and the depth error (mean squared error).}
\label{tab:pixerror}
\end{table*}

\end{document}